\documentclass[10pt,twocolumn,letterpaper]{article}

\usepackage{cvpr}
\usepackage{times}
\usepackage{epsfig}
\usepackage{graphicx}
\usepackage{amsmath}
\usepackage{amssymb}

\usepackage{graphicx}
\usepackage{subcaption}
\usepackage[table]{xcolor}
\usepackage{url}
\usepackage{algorithm}
\usepackage[noend]{algpseudocode}
\usepackage{xspace}
\usepackage{booktabs}
\usepackage{multirow}
\usepackage{siunitx}

\usepackage[pagebackref=true,breaklinks=true,letterpaper=true,colorlinks,bookmarks=false]{hyperref}

\cvprfinalcopy 

\def\cvprPaperID{2828} 

\ifcvprfinal\fi
\begin{document}

\title{Human Semantic Parsing for Person Re-identification}

\author{
\begin{tabular}{cccc}
Mahdi M. Kalayeh\textsuperscript{1*} &
Emrah Basaran\textsuperscript{2*} &
Muhittin G\"{o}kmen\textsuperscript{3}\\
{\tt\small mahdi@eecs.ucf.edu} &
{\tt\small basaranemrah@itu.edu.tr} &
{\tt\small gokmenm@mef.edu.tr}
\end{tabular}\\
\begin{tabular}{cc}
Mustafa E. Kamasak\textsuperscript{2} &
Mubarak Shah\textsuperscript{1}\\
{\tt\small kamasak@itu.edu.tr} &
{\tt\small shah@crcv.ucf.edu}
\end{tabular}\\\\
\begin{tabular}{c}
\textsuperscript{1}Center for Research in Computer Vision, University of Central Florida\\
\textsuperscript{2}Dept. of Computer Engineering, Istanbul Technical University\\
\textsuperscript{3}Dept. of Computer Engineering, MEF University\\
\end{tabular}
}


\maketitle

\begin{abstract}
Person re-identification is a challenging task mainly due to factors such as background clutter, pose, illumination and camera point of view variations. These elements hinder the process of extracting robust and discriminative representations, hence preventing different identities from being successfully distinguished. To improve the representation learning, usually local features from human body parts are extracted. However, the common practice for such a process has been based on bounding box part detection. In this paper, we propose to adopt human semantic parsing which, due to its pixel-level accuracy and capability of modeling arbitrary contours, is naturally a better alternative. Our proposed SPReID integrates human semantic parsing in person re-identification and not only considerably outperforms its counter baseline, but achieves state-of-the-art performance. We also show that, by employing a \textit{simple} yet effective training strategy, standard popular deep convolutional architectures such as Inception-V3 and ResNet-152, with no modification, while operating solely on full image, can dramatically outperform current state-of-the-art. Our proposed methods improve state-of-the-art person re-identification on: Market-1501 \cite{zheng2015scalable} by $\sim$\textbf{17}\% in mAP and $\sim$\textbf{6}\% in rank-1, CUHK03 \cite{li2014deepreid} by $\sim$\textbf{4}\% in rank-1 and DukeMTMC-reID \cite{zheng2017unlabeled} by $\sim$\textbf{24}\% in mAP and $\sim$\textbf{10}\% in rank-1.
\end{abstract}
\let\thefootnote\relax\footnote{\textsuperscript{*}Authors contributed equally}
\section{Introduction}
Given a query image, person re-identification is the problem of retrieving all the images of the same identity from a large gallery, where query and gallery images are captured by distinctively different cameras which may or may not have any field-of-view overlap. Hence it can be seen as a cross-camera data association problem. 

Person re-identification is a very challenging task. First, when a single person is captured by two different cameras, the illumination conditions, background clutter, occlusion, observable human body parts, and perceived posture of the person can be dramatically different. Second, even within a single camera, the aforementioned conditions can vary through time as the person moves and engages in different actions (\textit{e.g} suddenly taking something out of a bag while walking). Third, gallery itself usually consists of diverse images of a single person from multiple cameras, which given the above factors, generates a huge intra-class variation impeding the generalization of the learned representations. Fourth, compared to problems such as object recognition or detection, images in person re-identification benchmarks are usually of lower resolution, making it difficult to extract distinctive attributes to distinguish one identity from another. Considering the above challenges, an effective person re-identification system is obliged to learn representations that are identity-specific, context-invariant and agnostic with respect to the camera point of view.

In recent past, improving global (image-level) representation by leveraging local (part-level) features extracted from human body parts has been the main theme of person re-identification research. While an image-level representation is prone to background clutter and occlusion, part-level representations are supposed to be more robust. However, part detection in low resolution images has its very own challenges and any error in that stage can propagate to the entire person re-identification system. That is why some research works prefer to simply extract representations from multiple image patches, often horizontal strips, that are loosely associated to human body parts. On the other hand, almost all of the previous works which involve body parts begin with an often off-the-shelf pose estimation model and infer corresponding bounding boxes from predicted joint locations. The person re-identification systems then process the global and local representations in what can be coarsely seen as multi-branch deep convolutional neural network (CNN) architectures. These models while delivering very good results, usually consist of many sub-models that are trained in multiple stages, tailored specifically for person re-identification problem. By studying recent literature, we raise two major questions, in this paper. \textbf{First}, are such \textit{complex} models necessary to improve the performance of person re-identification? \textbf{Second}, are the local features best captured using bounding boxes on human body parts?

Addressing the first question, we show that a \textit{simple} model based on Inception-V3 \cite{szegedy2016rethinking} with no bells and whistles, operating solely on full body images and optimized in a straightforward training procedure can outperform current state-of-the art. Unlike recent research works which commonly adopt, binary or triplet losses, we train our model using softmax cross-entropy at two different input resolutions. Using re-ranking as a post processing technique, the improvement margin further increases. 

To address the second question, we propose using semantic segmentation, more specifically human semantic parsing, as an alternative to bounding boxes in order to extract local features from human body parts. While bounding boxes are coarse, can include background, and cannot capture deformable nature of human body, semantic segmentation is able to precisely localize arbitrary contours of various body parts even under severe pose variations. We begin by training a human semantic parsing model that learns to segment human body into multiple semantic regions and then use them to exploit local cues for person re-identification. We analyze two variations for integrating human semantic parsing into re-identification and show that they provide complementary representations. The contributions of this paper are as follows:

\begin{itemize}
    \item Through extensive set of experiments, we show that, our \textit{simple} yet effective training procedure can significantly outperform current state-of-the-art. We verify our observations using two standard deep convolutional architectures, namely Inception-V3 \cite{szegedy2016rethinking} and ResNet-152 \cite{he2016deep} on three different benchmarks.
    
    \item We propose SPReID, where human semantic parsing is employed to harness local visual cues for person re-identification. To do so, we train our very own semantic segmentation model and show that it not only helps improving person re-identification, but also achieves state-of-the-art performance on human semantic parsing problem, demonstrating the quality of our model.
    
    \item We improve state-of-the-art person re-identification performance on: Market-1501 \cite{zheng2015scalable} by $\sim$\textbf{17}\% in mAP and $\sim$\textbf{6}\% in rank-1, CUHK03 \cite{li2014deepreid} by $\sim$\textbf{4}\% in rank-1 and DukeMTMC-reID \cite{zheng2017unlabeled} by $\sim$\textbf{24}\% in mAP and $\sim$\textbf{10}\% in rank-1.
\end{itemize}

The remainder of this paper is organized as follows. Section \ref{sec:related_work} offers a brief overview of the person re-identification literature. We then present our method in Section \ref{sec:methodology}. Experimental results are discussed in Section \ref{sec:experiments}, followed by the implementation details in Section \ref{sec:implementation_detail}. Finally, we conclude the paper in Section \ref{sec:conclusion}.

\section{Related Work}\label{sec:related_work}
In recent years, significant progress has been achieved in different computer vision areas, including in person re-identification, thanks to the emergence of deep learning, and in particular deep convolutional neural networks. Challenges due to variations in pose and illumination, occlusion and background clutter in the person re-identification problem have resulted in the research community to focus on two major sub-problems, namely feature representation and similarity or distance metrics. Improvements in feature representation have mainly been achieved by leveraging local cues while in the latter, similarity measures such as contrastive or triplet loss have been studied. Next, we briefly survey the person re-identification literature.

To obtain robust representations, authors in \cite{li2017learning, zhao2017spindle} augment a global representation by employing human body parts. Specifically, Li \textit{et al.} \cite{li2017learning} learn the body parts roughly as head-shoulder, upper-body and lower-body using a spatial transformer network \cite{jaderberg2015spatial}. Then multiple streams, with shared weights, through a multi-scale CNN structure process these parts and ultimately concatenate them with a global representation. Zhao \textit{et al.} \cite{zhao2017spindle} use region proposal network, trained on an auxiliary pose dataset, to detect body parts. Part representations are then gradually combined and finally fused with the global representation. Their proposed model is very complex and is trained through multiple non-trivial stages. While avoiding to explicitly detect human body parts, authors in \cite{zhu2017part, cheng2016person} try to benefit from local cues by extracting multiple patches from image which are loosely associated to human body parts. Such frameworks cannot address the part misalignment properly. Taking a slightly different approach, in \cite{liu2017end, rahimpour2017person}, authors develop attention-based models where respectively, a Long Short-Term Memory (LSTM) \cite{hochreiter1997long} and a gradient-based attention model dynamically focus on distinctive regions in the image. Some works \cite{zheng2017pose, su2017pose} have tried to address the misalignment issue by explicitly integrating pose estimation into person re-identification where off-the-shelf pose estimation models are used to initialize part locations as quadrilaterals which then are aligned via affine transformation or spatial transformer network \cite{jaderberg2015spatial}.

Also, there has been some attempts \cite{schumann2017person, su2017multi} to improve person re-identification performance using person attributes. These attributes usually contain high-level semantic information that are supposedly invariant to pose, illumination and camera point of view. However, one should note that the exact same conditions make reliable detection of those attributes very challenging.

Several loss functions have been adopted for person re-identification. Some like \cite{varior2016gated, varior2016siamese, li2014deepreid, ahmed2015improved} have used positive and negative image pairs through contrastive and binary (verification) losses to train their neural network models. Others \cite{ding2015deep, chen2017beyond, cheng2016person} have employed triplet loss which requires a tuple of anchor, positive and negative images where the training objective is to simultaneously pushing the positive image towards the anchor while pulling the negative image away from it. These loss functions are very suitable for person re-identification due to its retrieval nature. However, their effectiveness is highly dependent on how the training pairs/triplets are chosen. Easy to distinguish pairs/triplets do not help the learning since no error signal will be backpropagated while the hard ones can result in the training process to diverge.
Unlike the aforementioned approaches, Zhao \textit{et al.} \cite{zhao2017spindle}, adopt simple multi-class classification loss while \cite{zheng2016discriminatively, mclaughlin2016recurrent} use a combination of both classification and verification losses.

In contrast to the above works, we propose to employ human semantic parsing to extract local regions from human body. We argue that semantic segmentation, due to its pixel-level accuracy, is naturally more suitable than bounding box part localization to cope with person re-identification challenges. To the best of our knowledge, we are the first to propose the integration of human semantic parsing into person re-identification.

\section{Methodology}\label{sec:methodology}
In this work, unless specified otherwise, we use Inception-V3 \cite{szegedy2016rethinking} as the CNN backbone for both human semantic parsing and person re-identification models. Therefore, we begin by briefly describing the Inception-V3 \cite{szegedy2016rethinking} architecture. Then, we provide details for our human semantic parsing model and finally explain how to integrate it into our proposed person re-identification framework.

\subsection{Inception-V3 Architecture}
Inception-V3 \cite{szegedy2016rethinking} is a 48-layers deep convolutional architecture. Since it employs global average pooling instead of fully-connected layer, it can operate on arbitrary input image sizes. While being shallower than different variations of popular ResNet \cite{he2016deep}, our experiments show that it gives competitive and in cases even better results than ResNet-152 \cite{he2016deep}, while being dramatically less computationally expensive. We will provide quantitative comparison between different choices of the backbone architecture.

The Inception-V3 \cite{szegedy2016rethinking} has an output stride of 32, where the activation size quickly reduces to $\frac{1}{8}$ of the input image resolution within the first seven layers. Such reduction is achieved by two convolution and one max pooling layer that operate with the stride of 2. The network follows by three blocks of Inception layers separated by two grid reduction modules. Spatial resolution of the activations remains intact within the Inception blocks, while grid reduction modules halve the activation size and increase the number of channels. Then, the output of the last Inception block is aggregated via global average pooling to produce a 2048-D feature vector. For more details on the architecture, readers are encouraged to refer to \cite{szegedy2016rethinking}.

\subsection{Human Semantic Parsing Model}\label{subsec:parsing_model}
In order to exploit local cues for person re-identification, we propose to employ human semantic parsing. We argue that semantic segmentation due to its pixel-level accuracy and robustness to pose variation is naturally superior to bounding box part detection.

We use Inception-V3 \cite{szegedy2016rethinking} as the backbone architecture of our human semantic parsing model. However, we make two modifications to adopt it for the semantic segmentation task. The quality of human semantic parsing heavily relies on the final activations to be of sufficient resolution. Hence, we change the stride of the last grid reduction module in the Inception-V3 \cite{szegedy2016rethinking} from 2 to 1 resulting in an output stride of 16 compared to 32 in the original architecture. To cope with the extra computation that consequently is added to the last Inception block, corresponding convolution filters are replaced with the dilated convolution \cite{yu2015multi}. We then remove the global average pooling and add an atrous spatial pyramid pooling (rates=3,6,9,12)\cite{chen2017rethinking} followed by a 1$\times$1 convolution layer as the classifier. This would allow us to perform multi-class classification in pixel-level and is a standard approach, commonly used in semantic segmentation architectures \cite{chen2016deeplab, chen2017rethinking}.

\begin{figure*}
    \centering
    \begin{subfigure}[b]{\textwidth}
        \includegraphics[width=\textwidth]{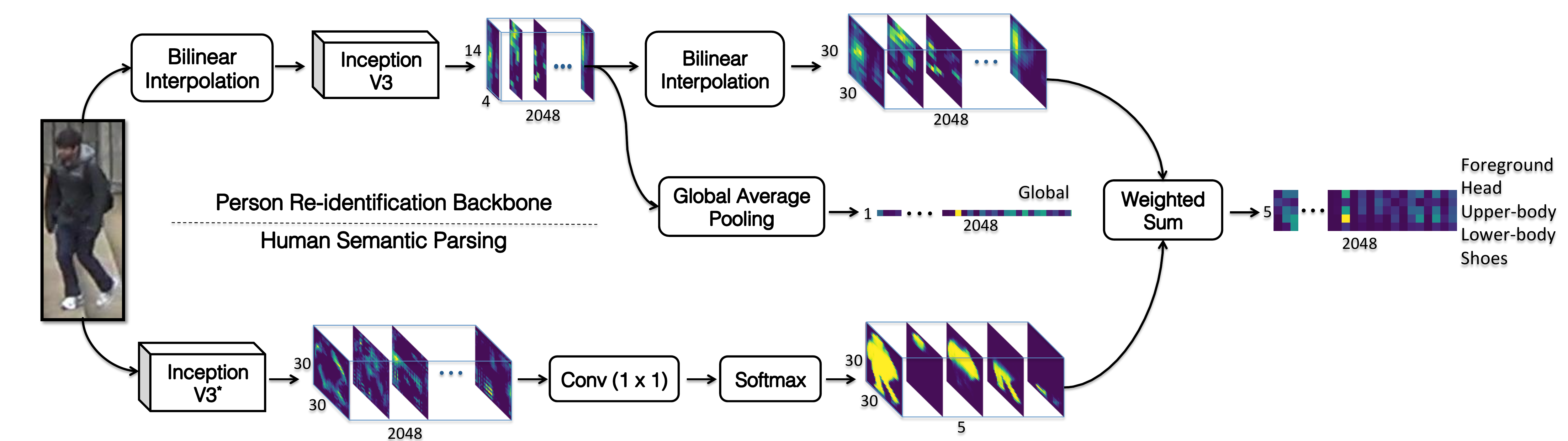}
    \end{subfigure}
    \caption{SPReID framework: our proposed person re-identification model first transforms the input RGB image into a tensor of activations via a convolutional backbone while simultaneously generating probability maps associated to different semantic regions of human body using the human semantic parsing branch. Note that the Inception-V3 module in the lower branch is denoted as Inception-V3\textsuperscript{*}. That refers to the modifications which we applied (ref. Section \ref{subsec:parsing_model}) to the original Inception-V3 \cite{szegedy2016rethinking} architecture. SPReID then uses the aforementioned probability maps to aggregate the convolutional activations from different semantic regions of human body.}\label{fig:method}
\end{figure*}

\subsection{Person Re-identification Model}
Our person re-identification model, illustrated in Figure \ref{fig:method} consists of a convolutional backbone, a human semantic parsing branch and two aggregation heads. From now on, we refer to it as \textbf{SPReID}: Human \textbf{S}emantic \textbf{P}arsing for Person \textbf{Re}-\textbf{id}entification. The person re-identification backbone in SPReID is exactly Inception-V3 \cite{szegedy2016rethinking} with a minor modification of removing global average pooling layer. Hence, it generates a tensor of 2048 channels with the output stride of 32.

Our baseline person re-identification model simply aggregates the output activations of the convolutional backbone using global average pooling. Corresponding aggregation head, shown in Figure \ref{fig:method} generates a 2048-D global representation. To train the network, we pass it to a multi-class classification (over different identities) objective with softmax cross-entropy loss. To avoid clutter, we are not showing the loss in Figure \ref{fig:method}. At the test time, final representations before the classifier layer are used to retrieve correct matches of a given query from the gallery. In Section \ref{sec:experiments}, we show how the performance of the baseline model varies if we change the backbone architecture from Inception-V3 \cite{szegedy2016rethinking} to ResNet-50 \cite{he2016deep} and ResNet-152 \cite{he2016deep}.

To exploit the local visual cues, we use the probability maps associated to five different body regions, namely foreground, head, upper-body, lower-body and shoes. These probability maps are generated by the human semantic parsing model and are $\ell_{1}$-normalized per channel. In SPReID, we pool the output activations of the CNN backbone multiple times, each time using one of the five probability maps. This is in contrast with global average pooling, which is agnostic with respect to where in the spatial domain activations occur. It is not hard to see that exclusively pooling activations within different semantic regions associated to human body parts can be seen as a weighted sum operation where the probability maps are used as weights. From an implementation point of view, this is equal to a matrix multiplication between the output of re-identification backbone and human semantic parsing where their corresponding spatial domain is flattened. Such a procedure results in five 2048-D feature vectors each exclusively representing one human body region. Next, we perform element-wise max operation over representations of head, upper-body, lower-body and shoes and concatenate the outcome with the foreground and previously described global representation from the full image. Our proposed technique is applicable to any convolutional backbone choice and adds minimal computation to the naive global average pooling which serves as our baseline person re-identification model. Note that since the human semantic parsing model usually operates on higher resolution images, the re-identification backbone, as shown in Figure \ref{fig:method} uses bilinear interpolation to initially scale down the input images and then scale up the final activations to match the ones in human semantic parsing branch.

\section{Experiments}\label{sec:experiments}

\subsection{Datasets and Evaluation Measures}\label{sec:datasets}
To evaluate our proposed methods, we use three publicly available large-scale person re-identification benchmarks namely Market-1501 \cite{zheng2015scalable}, CUHK03 \cite{li2014deepreid} and DukeMTMC-reID \cite{zheng2017unlabeled}. Market-1501 \cite{zheng2015scalable} dataset consists of 32,668 images of 1,501 subjects captured by 5 high-resolution and one low-resolution camera.  In this dataset, to obtain the person bounding boxes, Deformable Part Model (DPM) \cite{felzenszwalb2010object} is used. Therefore, there are misaligned detected boxes within the dataset. In its standard evaluation protocol, the training set consists of 751 identities and has a total of 12,936 images. In the test set, images of 750 identities which have not appeared in the training are used to create gallery and query sets. These sets respectively contain 19,734 and 3,368 images.

DukeMTMC-reID \cite{zheng2017unlabeled} dataset consists of the person images which are extracted from the DukeMTMC \cite{ristani2016MTMC} tracking dataset. DukeMTMC contains images taken from 8 high-resolution cameras, and person bounding boxes are hand-annotated. The standard evaluation protocol \cite{zheng2017unlabeled} of DukeMTMC-reID dataset is in the same format as Market-1501. Specifically, 16,522 images of 702 persons are reserved as training set. For gallery and probe, respectively 16,522 and 2,228 images associated to 702 identities that do not appear in the training set are used. In CUHK03 \cite{li2014deepreid} dataset, there are 13,164 images with a total of 1,467 identities. These images were recorded by 6 surveillance cameras and each person is viewed by 2 different cameras. For the experiments conducted on this benchmark, both manually annotated and DPM-detected bounding boxes can be used. The evaluation protocol of CUHK03 is in a different format than the other two datasets. In our experiments, we are following the standard protocol detailed in \cite{li2014deepreid} and reporting the results on the manually annotated images.

In addition to these datasets that were used in evaluation, we utilize 3DPeS \cite{baltieri20113dpes}, CUHK01 \cite{li2012human}, CUHK02 \cite{li2013locally}, PRID \cite{hirzer11}, PSDB \cite{xiao2016end}, Shinpuhkan \cite{kawanishi2014shinpuhkan2014} and VIPeR \cite{gray2007evaluating} datasets to augment our training data. The training splits of these datasets, in addition to Market-1501 \cite{zheng2015scalable}, CUHK03 \cite{li2014deepreid} and DukeMTMC-reID \cite{zheng2017unlabeled}, are aggregated to create a large training set which consists of $\sim$111,000 images. We evaluate the quality of different person re-identification models using Cumulative Matching Characteristic (CMC) curves and mean average precision (mAP). All the experiments are performed in single query setting.

\subsection{Training the Networks}\label{sec:training_networks}
To train our person re-identification models, we aggregate 10 different person re-identification benchmarks, detailed in Section \ref{sec:datasets}, which results in a total of $\sim$111,000 images of $\sim$17,000 identities. The baseline models solely operate on full image with no use of semantic segmentation. We begin by training them for 200K iterations using input images of size 492$\times$164. Then, we fine-tune each one for an additional 50K iteration but on higher input resolution of 748$\times$246. Fine-tuning is conducted on Market-1501, CUHK03 and DukeMTMC-reID datasets separately. Training of SPReID is done on the aggregation of 10 datasets with the exact same setting as above. The input image resolution in its associated experiments is set to 512$\times$170.

\begin{figure}
    \centering
    \begin{subfigure}[b]{0.48\textwidth}
        \includegraphics[width=\textwidth]{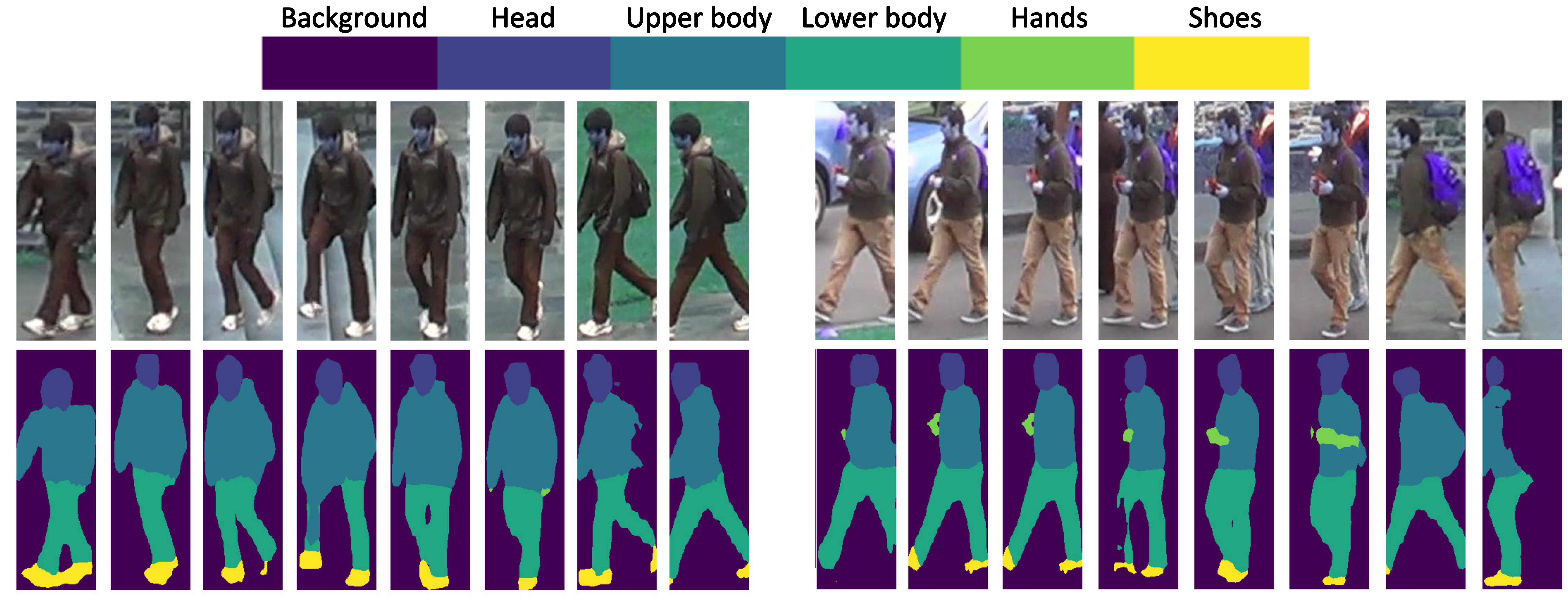}
    \end{subfigure}
    \caption{Examples of the segmentation masks generated by our human semantic parsing model on random images from DukeMTMC-reID \cite{zheng2017unlabeled} person re-identification benchmark.}\label{fig:segteaser}
\end{figure}

We train the human semantic parsing model on Look into Person (LIP) \cite{gong2017look} dataset which consists of $\sim$30,000 images with 20 semantic labels\textsuperscript{1}\footnote{\textsuperscript{1}Background, Hat, Hair, Glove, Sunglasses, Upper-clothes, Dress, Coat, Socks, Pants, Jumpsuits, Scarf, Skirt, Face, Right-arm, Left-arm, Right-leg, Left-leg, Right-shoe and Left-shoe}. The probability of predictions for different regions are then grouped together to create 5 coarse labels\textsuperscript{2}\footnote{\textsuperscript{2}Foreground, Head, Upper-body, Lower-body and Shoes} in order to parse human body for the person re-identification. Our experiments indicate that the human semantic parsing model is capable of decently localizing various human body parts even under severe pose variation and occlusion. Despite being out of the scope of this work, to demonstrate the quality of our human semantic parsing, we show in Table \ref{tab:results_on_LIP} that, on the validation set of LIP \cite{gong2017look}, our model outperforms the current state-of-the-art. Figure \ref{fig:segteaser} illustrates how our human semantic parsing model segments example images from DukeMTMC-reID \cite{zheng2017unlabeled} person re-identification benchmark.

\begin{table}
	\centering
	\tabcolsep=0.07cm
	\begin{tabular}{lccc}
		\toprule
		method & overall acc. & mean acc. & mean IoU\\
		\midrule
		SegNet \cite{badrinarayanan2015segnet} & 69.04 & 24.00 & 18.17\\
        FCN-8s \cite{long2015fully} & 76.06 & 36.75 & 28.29\\
        DeepLabV2 \cite{chen2016deeplab} & 82.66 & 51.64 & 41.64\\
        Attention \cite{chen2016attention} & 83.43 & 54.39 &  42.92\\
        DeepLabV2 + SSL \cite{gong2017look} & 83.16 & 52.55 & 42.44\\
        Attention + SSL \cite{gong2017look} & 84.36 & 54.94 & 44.73\\
        Ours & \textbf{85.07} & \textbf{60.54} & \textbf{48.16}\\
		\bottomrule\\
	\end{tabular}
	\caption{Performance (\%) comparison of human semantic parsing on the validation split of LIP \cite{gong2017look}.}
	\label{tab:results_on_LIP}
\end{table}

\subsection{Person Re-identification Performance}

In this section, we begin by analyzing the performance of our baseline person re-identification models. We will show the effect of input image resolution, fine-tuning on large image size, different choices for the re-identification backbone, and finally weight sharing among aggregation heads. We show that the baseline models, thanks to our simple yet well designed training strategy, can outperform the current state-of-the-art with large margin. Then, we quantitatively illustrate the effectiveness of SPReID in harnessing human semantic parsing for person re-identification. We conclude this section by comparison with the state-of-the-art person re-identification on three large-scale benchmarks.

\textbf{Effect of input image resolution:} In Table \ref{tab:results_inputres}, we show quantitative results from our Inception-V3 baseline model when different input resolutions are used to train the network. Other than that, the rest of settings/parameters are the same for all models. We observe that on all three datasets, training on higher resolution input images yields a better performance measured by either mAP or re-identification rate. Though such gap tends to shrink when we consider rank-10 versus rank-1, as it is expected. Model-S, Model-M and Model-L are trained on $\sim$111K images of $\sim$17K identities when we merge 10 different person re-identification datasets. Since training on high resolution images is computationally expensive, in order to further push the performance boundaries, we take a trained Model-L and fine-tune it with input images of 748$\times$246 which is $\sim$1.5 times larger than what Model-L has been originally trained with. Table \ref{tab:results_inputres} shows that such a fine-tuning practice, denoted as Model-L\textsuperscript{\textit{ft}}, yields an average of \textbf{4.75}\% mAP, and \textbf{1.71}\% rank-1 score on the top of Model-L. Hence, we confirm the advantages of training person re-identification models using large input image sizes.

\begin{table}
	\centering
	\begin{tabular}{lcccc}
		\toprule
		\multicolumn{5}{c}{\textbf{Market-1501}}\\
		\midrule
		model & input size & mAP(\%) & rank-1 & rank-10\\
		\midrule
		Model-S & 246$\times$82 & 64.45 & 84.06 & 95.55\\
		Model-M & 375$\times$125 & 72.14 & 88.18 & 96.64\\
		Model-L & 492$\times$164 & 73.06 & 88.87 & 97.00\\
		Model-L$\textsuperscript{\textit{ft}}$ & 748$\times$246 & \textbf{76.56} & \textbf{90.8} & \textbf{97.71}\\
		\toprule
		\multicolumn{5}{c}{\textbf{CUHK03}}\\
		\midrule
		model & input size & mAP(\%) & rank-1 & rank-10\\
		\midrule
		Model-S & 246$\times$82 & -- & 81.78 & 98.12\\
		Model-M & 375$\times$125 & -- & 85.66 & 98.90\\
		Model-L & 492$\times$164 & -- & 87.91 & 98.41\\
		Model-L$\textsuperscript{\textit{ft}}$ & 748$\times$246 & -- & \textbf{88.73} & \textbf{98.94}\\
		\toprule
		\multicolumn{5}{c}{\textbf{DukeMTMC-reID}}\\
		\midrule
		model & input size & mAP(\%) & rank-1 & rank-10\\
		\midrule
		Model-S & 246$\times$82 & 53.73 & 74.87 & 88.51\\
		Model-M & 375$\times$125 & 59.98 & 79.85 & 90.89\\
		Model-L & 492$\times$164 & 59.87 & 79.08 & 90.26\\
		Model-L$\textsuperscript{\textit{ft}}$ & 748$\times$246 & \textbf{63.27} & \textbf{80.48} & \textbf{91.65}\\
		\bottomrule\\
	\end{tabular}
	\caption{Effect of input image resolution on Inception-V3 baseline model, measured by mAP and re-identification rate. We observe that higher input image resolution and fine-tuning can provide considerable performance gain. Small, medium and large models are respectively indicated as Model-S, Model-M and Model-L.}
	\label{tab:results_inputres}
\end{table}

\textbf{Choice of re-identification backbone architecture:} Table \ref{tab:results_architecture} shows the effect of varying the re-identification backbone architecture in our baseline model. Inception-V3 \cite{szegedy2016rethinking} despite its considerably shallower architecture, provides a very competitive performance with ResNet-152 \cite{he2016deep}, while significantly outperforming ResNet-50 \cite{he2016deep}, which is of approximately the same depth. Table \ref{tab:results_architecture} also shows that the performance gain achieved by fine-tuning on high resolution images (ref. Table \ref{tab:results_inputres}) is valid across variety of the architecture choices. In our experiments, we observe that ResNet-152 is 3 times more computationally expensive (measured by forward+backward time) than Inception-V3. Hence, given their relatively similar performance, we chose Inception-V3 as our main backbone architecture.

\begin{table}
	\centering
	\begin{tabular}{lccc}
		\toprule
		\multicolumn{4}{c}{\textbf{Market-1501}}\\
		\midrule
		model & mAP(\%) & rank-1 & rank-10\\
		\midrule
		Inception-V3 & \textbf{73.06} & \textbf{88.87} & \textbf{97.00}\\
		ResNet-50 & 66.32 & 85.10 & 95.75\\
		ResNet-152 & 72.95 & 88.33 & 96.88\\
		\midrule
		Inception-V3$\textsuperscript{\textit{ft}}$ & 76.56 & \textbf{90.80} & \textbf{97.71}\\
		ResNet-50$\textsuperscript{\textit{ft}}$ & 72.97 & 87.92 & 96.76\\
		ResNet-152$\textsuperscript{\textit{ft}}$ & \textbf{77.96} & 90.71 & 97.65\\
		\toprule
		\multicolumn{4}{c}{\textbf{CUHK03}}\\
		\midrule
		model & mAP(\%) & rank-1 & rank-10\\
		\midrule
		Inception-V3 & -- & 87.91 & 98.41\\
		ResNet-50 & -- & 85.88 & 99.19\\
		ResNet-152 & -- & \textbf{88.01} & \textbf{99.27}\\
		\midrule
		Inception-V3$\textsuperscript{\textit{ft}}$ & -- & 88.73 & 98.94\\
		ResNet-50$\textsuperscript{\textit{ft}}$ & -- & 89.08 & 99.15\\
		ResNet-152$\textsuperscript{\textit{ft}}$ & -- & \textbf{90.38} & \textbf{99.46}\\
		\toprule
		\multicolumn{4}{c}{\textbf{DukeMTMC-reID}}\\
		\midrule
		model & mAP(\%) & rank-1 & rank-10\\
		\midrule
		Inception-V3 & 59.87 & 79.08 & 90.26\\
		ResNet-50 & 54.77 & 73.70 & 88.02\\
		ResNet-152 & \textbf{62.42} & \textbf{79.62} & \textbf{90.80}\\
		\midrule
		Inception-V3$\textsuperscript{\textit{ft}}$ & 63.27 & 80.48 & 91.65\\
        ResNet-50$\textsuperscript{\textit{ft}}$ & 59.72 & 77.74 & 90.84\\
		ResNet-152$\textsuperscript{\textit{ft}}$ & \textbf{67.02} & \textbf{83.26} & \textbf{92.95}\\
		\bottomrule\\
	\end{tabular}
	\caption{Effect of backbone architecture in our baseline person re-identification model, measured by mAP and re-identification rate.}
	\label{tab:results_architecture}
\end{table}

\textbf{SPReID Performance:}
Table \ref{tab:results_segmentation} compares the performance of our proposed SPReID against the Inception-V3 baseline person re-identification. All the models are trained using the settings detailed in Section \ref{sec:training_networks}. We observe that both with and without foreground variations, respectively denoted as SPReID$\textsuperscript{\textit{w/fg}}$ and SPReID$\textsuperscript{\textit{wo/fg}}$, outperform Inception-V3 baseline while their combination ($\ell_{2}$-normalization+concatenation) results in further performance gains. Exploiting human semantic parsing through SPReID improves the baseline re-identification model on: Market-1501 \cite{zheng2015scalable} by \textbf{6.61}\% in mAP and \textbf{2.58}\% in rank-1, CUHK03 \cite{li2014deepreid} by \textbf{3.33}\% in rank-1 and DukeMTMC-reID \cite{zheng2017unlabeled} by \textbf{8.91}\% in mAP and \textbf{4.22}\% in rank-1. Since the only difference between Inception-V3 baseline and SPReID is in how they aggregate the activations of the final convolution layer, we can confirm the advantage of our proposed method in effectively harnessing human semantic parsing to improve person re-identification.

\begin{table}
	\centering
	\begin{tabular}{lccc}
		\toprule
		\multicolumn{4}{c}{\textbf{Market-1501}}\\
		\midrule
		model & mAP(\%) & rank-1 & rank-10\\
		\midrule
		Inception-V3 & 73.06 & 88.87 & 97.00\\
		SPReID$\textsuperscript{\textit{w/fg}}$ & 78.66 & 90.97 & 97.71\\
		SPReID$\textsuperscript{\textit{wo/fg}}$ & 78.06 & 90.74 & 97.80\\
		SPReID$\textsuperscript{\textit{combined}}$ & \textbf{79.67} & \textbf{91.45} & \textbf{98.1}\\
		\toprule
		\multicolumn{4}{c}{\textbf{CUHK03}}\\
		\midrule
		model & mAP(\%) & rank-1 & rank-10\\
		\midrule
		Inception-V3 & -- & 87.91 & 98.41\\
		SPReID$\textsuperscript{\textit{w/fg}}$ & -- & 89.57 & 99.19\\
		SPReID$\textsuperscript{\textit{wo/fg}}$ & -- & \textbf{91.29} & 98.93\\
		SPReID$\textsuperscript{\textit{combined}}$ & -- & 91.21 & \textbf{99.2}\\
		\toprule
		\multicolumn{4}{c}{\textbf{DukeMTMC-reID}}\\
		\midrule
		model & mAP(\%) & rank-1 & rank-10\\
		\midrule
		Inception-V3 & 59.87 & 79.08 & 90.26\\
		SPReID$\textsuperscript{\textit{w/fg}}$ & 67.20 & 82.32 & 92.32\\
		SPReID$\textsuperscript{\textit{wo/fg}}$ & 67.11 & 82.14 & 92.24\\
        SPReID$\textsuperscript{\textit{combined}}$ & \textbf{68.78} & \textbf{83.3} & \textbf{92.91}\\
		\bottomrule\\
	\end{tabular}
	\caption{Effect of utilizing human semantic parsing by SPReID to improve Inception-V3 person re-identification baseline, measured by mAP and re-identification rate.}
	\label{tab:results_segmentation}
\end{table}

\textbf{Effect of weight sharing:}
SPReID model illustrated in Figure \ref{fig:method} has two aggregation heads. One simply performs global average pooling while the other uses probability maps associated to different human body parts as weights to aggregate convolutional activations. Table \ref{tab:results_weightsharing} compares two scenarios based on whether or not the two aggregation heads share the re-identification backbone. We observe that while exclusive backbone achieves slightly better results than weight sharing, with the exception of CUHK03 \cite{li2014deepreid} the margin shrinks after fine-tuning on very high image resolutions. It is worth noting that in both scenarios, SPReID outperforms Inception-V3 baseline (ref. Table \ref{tab:results_segmentation}).

\begin{table}
	\centering
	\begin{tabular}{lccc}
		\toprule
		\multicolumn{4}{c}{\textbf{Market-1501}}\\
		\midrule
		model & weight sharing & mAP(\%) & rank-1\\
		\midrule
		SPReID$\textsuperscript{\textit{w/fg}}$ & N & 78.66 & 90.97\\
		SPReID$\textsuperscript{\textit{w/fg}}$ & Y & 77.62 & 90.88\\
		SPReID$\textsuperscript{\textit{w/fg-ft}}$ & N & \textbf{80.68} & \textbf{92.40}\\
		SPReID$\textsuperscript{\textit{w/fg-ft}}$ & Y & 80.54 & 92.34\\

		\toprule
		\multicolumn{4}{c}{\textbf{CUHK03}}\\
		\midrule
		model & weight sharing & mAP(\%) & rank-1\\
		\midrule
		SPReID$\textsuperscript{\textit{w/fg}}$ & N & -- & 89.57\\
		SPReID$\textsuperscript{\textit{w/fg}}$ & Y & -- & 87.69\\
		SPReID$\textsuperscript{\textit{w/fg-ft}}$ & N & -- & \textbf{92.57}\\
		SPReID$\textsuperscript{\textit{w/fg-ft}}$ & Y & -- & 89.68\\
		\toprule
		\multicolumn{4}{c}{\textbf{DukeMTMC-reID}}\\
		\midrule
		model & weight sharing & mAP(\%) & rank-1\\
		\midrule
		SPReID$\textsuperscript{\textit{w/fg}}$ & N & 67.20 & 82.32\\
		SPReID$\textsuperscript{\textit{w/fg}}$ & Y & 65.66 & 81.73\\
		SPReID$\textsuperscript{\textit{w/fg-ft}}$ & N & \textbf{69.79} & \textbf{84.02}\\
		SPReID$\textsuperscript{\textit{w/fg-ft}}$ & Y & 69.29 & 83.80\\
		\bottomrule\\
	\end{tabular}
	\caption{Effect of weight sharing in Inception-V3 backbone between global average pooling, and semantic based pooling branches of SPReID person re-identification architecture.}
	\label{tab:results_weightsharing}
\end{table}

\subsection{Comparison with the state-of-the-art}
Table \ref{tab:results_soa} shows the performance of our person re-identification models against the current state-of-the-art. For each dataset, the corresponding results are divided into three blocks, first one shows the performance of current state-of-the-art methods. Second block shows the performance of our baseline models with no human semantic parsing cues but trained using our two-stage training procedure. The third block shows the performance of SPReID.

From Table \ref{tab:results_soa}, we observe that the baseline person re-identification models when trained using our proposed training procedure outperform the current state-of-the-art. These results are particularly interesting, since the models are less complex and are also trained in a straightforward fashion. When utilizing re-ranking \cite{zhong2017re}, the improvement margin further increases. Therefore, we confirm that a simple model with no bells and whistles is sufficient to achieve state-of-the-art person re-identification performance. Table \ref{tab:results_soa} shows that SPReID can effectively harness local visual cues from human body parts. On all three datasets, SPReID\textsuperscript{\textit{combined-ft}} outperforms Inception-V3\textsuperscript{\textit{ft}} baseline with a large margin. Although, the gap reduces when models are combined with the strong ResNet-152\textsuperscript{\textit{ft}} baseline. Similar to the previous case, the performance will be further improved by employing re-ranking as post processing.

\begin{table}
	\centering
	\tabcolsep=0.15cm
	\begin{tabular}{lcccc}
		\toprule
		\multicolumn{5}{c}{\textbf{Market-1501}}\\
		\midrule
		method & mAP(\%) & rank-1 & rank-5 & rank-10\\
		\midrule
		Li \textit{et. al.} \cite{li2017learning} & 57.5 & 80.3 & -- & -- \\
		SVDNet \cite{sun2017svdnet} & 62.1 & 82.3 & 92.3 & 95.2\\
		DPAR \cite{zhao2017deeply} & 63.4 & 81.0 & 92.0 & 94.7\\
		JLML \cite{li2017person} & 65.5 & 85.1 & -- & -- \\
		Basel.+LSRO \cite{zheng2017unlabeled} & 66.1 & 84.0 & -- & --\\
		SSM \cite{bai2017scalable} & 68.8 & 82.2 & -- & --\\
		DaF \cite{yu2017divide} & 72.4 & 82.3 & -- & --\\
		Chen \textit{et. al.} \cite{chen2017person} & 73.1 & 88.9 & -- & --\\
		\midrule

		Inception-V3$\textsuperscript{\textit{ft}}$ & 76.56 & 90.8 & 96.35 & 97.71\\
		Inception-V3$\textsuperscript{\textit{ft*}}$ & 82.87 & 93.14 & 97.27 & 98.22\\
		+re-ranking\cite{zhong2017re} & 90.66 & 94.21 & 96.76 & 97.3\\
		\midrule
        SPReID$\textsuperscript{\textit{combined-ft}}$ & 81.34 & 92.54 & 97.15 & 98.1\\
        SPReID$\textsuperscript{\textit{combined-ft*}}$ & 83.36 & 93.68 & \textbf{97.57} & \textbf{98.4}\\
        +re-ranking\cite{zhong2017re} & \textbf{90.96} & \textbf{94.63} & 96.82 & 97.65\\

		\toprule
		\multicolumn{5}{c}{\textbf{CUHK03}}\\
		\midrule
		method & mAP(\%) & rank-1 & rank-5 & rank-10\\
		\midrule
		FT-JSTL+DGD \cite{xiao2016learning} & -- & 75.3 & -- & --\\
		SSM \cite{bai2017scalable} & -- & 76.6 & 94.6 & 98\\
		Spindle \cite{zhao2017spindle} & -- & 88.5 & 97.8 & 98.6\\
		DPAR \cite{zhao2017deeply} & -- & 85.4 & 97.6 & 99.4\\
		Chen \textit{et. al.} \cite{chen2017person} & 82.8 & 86.7 & -- & --\\
		HydraPlus \cite{liu2017hydraplus} & -- & 91.8 & 98.4 & 99.1 \\
		\midrule

		Inception-V3$\textsuperscript{\textit{ft}}$ & -- & 88.73 & 97.82 & 98.94\\
		Inception-V3$\textsuperscript{\textit{ft*}}$ & -- & 92.81 & 98.9 & 99.35\\
		+re-ranking\cite{zhong2017re} & -- & 95.18 & 99.18 & 99.6\\
		\midrule
		SPReID$\textsuperscript{\textit{combined-ft}}$ & -- & 93.89 & 98.76 & 99.51\\
		SPReID$\textsuperscript{\textit{combined-ft*}}$ & -- & 94.28 & 99.04 & 99.56\\
		+re-ranking\cite{zhong2017re} & -- & \textbf{96.22} & \textbf{99.34} & \textbf{99.7}\\

		\toprule
		\multicolumn{5}{c}{\textbf{DukeMTMC-reID}}\\
		\midrule
		method & mAP(\%) & rank-1 & rank-5 & rank-10\\
		\midrule
		Basel.+LSRO \cite{zheng2017unlabeled} & 47.1 & 67.7 & -- & --\\
		Basel.+OIM \cite{xiao2017joint} & -- & 68.1 & -- & --\\
		ACRN \cite{schumann2017person} & 52.0 & 72.6 & 84.8 & 88.9\\
		SVDNet \cite{sun2017svdnet} & 56.8 & 76.7 & 86.4 & 89.9\\
		Chen \textit{et. al.} \cite{chen2017person} & 60.6 & 79.2 & -- & --\\
		\midrule
		Inception-V3$\textsuperscript{\textit{ft}}$ & 63.27 & 80.48 & 88.78 & 91.65\\
		Inception-V3$\textsuperscript{\textit{ft*}}$ & 72 & 85.37 & 92.15 & 94.21\\
		+re-ranking\cite{zhong2017re} & 84.82 & \textbf{89.41} & 93.18 & \textbf{94.75}\\
		\midrule
        SPReID$\textsuperscript{\textit{combined-ft}}$ & 70.97 & 84.43 & 91.88 & 93.72\\
		SPReID$\textsuperscript{\textit{combined-ft*}}$ & 73.34 & 85.95 & 92.95 & 94.52\\
		+re-ranking\cite{zhong2017re} & \textbf{84.99} & 88.96 & \textbf{93.27} & \textbf{94.75}\\

		\bottomrule\\
	\end{tabular}
	\caption{Comparison with the state-of-the-art.\textsuperscript{*}indicates combination ($\ell_{2}$-normalization+concatenation) with ResNet-152$\textsuperscript{\textit{ft}}$.}
	\label{tab:results_soa}
\end{table}

\section{Implementation Details}\label{sec:implementation_detail}
\textbf{Person Re-identification:} In both training phases, mini-batch size is set to 15, momentum to 0.9 and we use weight decay and gradient clipping with 0.0005 and 2.0 for the respective values. Initial learning rate value is set to 0.01 in the first phase and reduces to 0.001 in the second phase. Throughout training, we decay the learning rate 10 times using exponential shift with the rate of 0.9. We train the models using Nesterov Accelarated Gradient \cite{bengio2013advances} and initialize the weights using pre-trained models on ImageNet \cite{russakovsky2015imagenet}. 

\textbf{Human Semantic Parsing:} We train our human semantic parsing model for 30K iterations where the initial learning rates for the Inception-V3 backbone, atrous spatial pyramid pooling and the 1$\times$1 convolution layer are respectively set to 0.01, 0.1 and 0.1. The rest of the parameters and settings are similar to the ones for person re-identification except the input resolution where 512$\times$512 input images are used.
\section{Conclusion}\label{sec:conclusion}
In this paper, we began by raising two major questions. First, whether to achieve state-of-the-art performance, the person re-identification models need to be \textit{complex}. Second, whether bounding boxes on human body parts is the best practice to harness local visual cues. Through this paper, we addressed both of these questions with extensive set of experiments. We showed that, indeed a \textit{simple} deep convolutional architecture when trained properly on large number of high resolution images can outperform the current state-of-the-art. We also demonstrated that, by exploiting human semantic parsing in our proposed SPReID framework, the performance of an state-of-the-art baseline model can be further improved. SPReID applies minimal modifications to the person re-identification backbone and offers a more natural solution for utilizing human body parts. We hope that, this work encourages the research community to invest more in employing human semantic parsing for person re-identification task.

\section*{Acknowledgments}
This research is based upon work supported in parts by the U. S. Army Research Laboratory and the U. S. Army Research Office (ARO) under contract/grant number  W911NF-14-1-0294; and the Office of the Director of National Intelligence (ODNI), Intelligence Advanced Research Projects Activity (IARPA), via IARPA R\&D Contract No. D17PC00345. The views and conclusions contained herein are those of the authors and should not be interpreted as necessarily representing the official policies or endorsements, either expressed or implied, of the ODNI, IARPA, or the U.S. Government. The U.S. Government is authorized to reproduce and distribute reprints for Governmental purposes notwithstanding any copyright annotation thereon. Emrah Basaran was supported by 2214-A programme of The Scientific and Technological Research Council of Turkey (T\"{U}B\.{I}TAK).

{\small
\bibliographystyle{ieee}
\bibliography{cvpr18_camera_ready.bib}
}

\end{document}